\documentclass[sigconf]{acmart}
\AtBeginDocument{%
  \providecommand\BibTeX{{%
    \normalfont B\kern-0.5em{\scshape i\kern-0.25em b}\kern-0.8em\TeX}}}

\usepackage{balance}
\usepackage{multirow}
\usepackage{hyperref}
\usepackage{enumitem}
\usepackage{bbding}
\usepackage{amsthm}
\copyrightyear{2022}
\acmYear{2022}
\setcopyright{acmlicensed}\acmConference[MM '22]{Proceedings of the 30th
ACM International Conference on Multimedia}{October 10--14, 2022}{Lisboa,
Portugal}
\acmBooktitle{Proceedings of the 30th ACM International Conference on
Multimedia (MM '22), October 10--14, 2022, Lisboa, Portugal}
\acmPrice{15.00}
\acmDOI{10.1145/3503161.3548265}
\acmISBN{978-1-4503-9203-7/22/10}

\begin{document}
%
%

\title{Delving Globally into Texture and Structure for Image Inpainting}

\author{Haipeng Liu}
\affiliation{%
\institution{School of Computer Science and Information Engineering, Hefei University of Technology}
\streetaddress{}
\city{Hefei}
\country{China}
\state{}}
\email{hpliu_hfut@hotmail.com}

\author{Yang Wang}
\authornote{Yang Wang is the corresponding author}
\affiliation{%
\institution{Key Laboratory of Knowledge Engineering with Big Data, Ministry of Education, Hefei University of Technology}
\streetaddress{}
\city{Hefei}
\country{China}
\state{}}
\email{yangwang@hfut.edu.cn }

\author{Meng Wang}
\affiliation{%
\institution{Key Laboratory of Knowledge Engineering with Big Data, Ministry of Education, Hefei University of Technology}
\streetaddress{}
\city{Hefei}
\country{China}
\state{}}
\email{eric.mengwang@gmail.com}

\author{Yong Rui}
\affiliation{%
\institution{Lenovo Research}
\streetaddress{}
\city{Beijing}
\country{China}}
\email{yongrui@lenovo.com}



\renewcommand{\shortauthors}{Haipeng Liu, Yang Wang, Meng Wang, \& Yong Rui}

\begin{abstract}
Image inpainting has achieved remarkable progress and inspired abundant methods, where the critical bottleneck is identified as how to fulfill the high-frequency structure and low-frequency texture information on the masked regions with semantics. To this end, deep models exhibit powerful superiority  to capture them, yet constrained on the local spatial regions. In this paper, we delve globally into  texture and structure information to well capture the semantics for image inpainting. As opposed  to the existing arts trapped on the independent local patches, the texture information of each patch is  reconstructed from all other patches across the whole image, to match the coarsely  filled information, especially the structure information over the masked regions. Unlike the current  decoder-only transformer within the pixel level for image inpainting, our model adopts the transformer  pipeline paired with both encoder and decoder. On one hand, the encoder captures the texture semantic  correlations of all patches across image via self-attention module. On the other hand, an adaptive patch  vocabulary is dynamically established in the decoder for the filled patches over the masked regions. Building on this,  a structure-texture matching attention module anchored on the known regions comes up to marry the  best of these two worlds for progressive inpainting via a probabilistic diffusion process. Our model is orthogonal to the fashionable  arts, such as Convolutional Neural Networks (CNNs), Attention and Transformer model, from the  perspective of texture and structure information for image inpainting. The extensive experiments over the benchmarks validate its superiority. Our code is available \href{https://github.com/htyjers/DGTS-Inpainting}{\textit{here}}.

\end{abstract}



\begin{teaserfigure}
\centering
  \includegraphics[width=15.7cm]{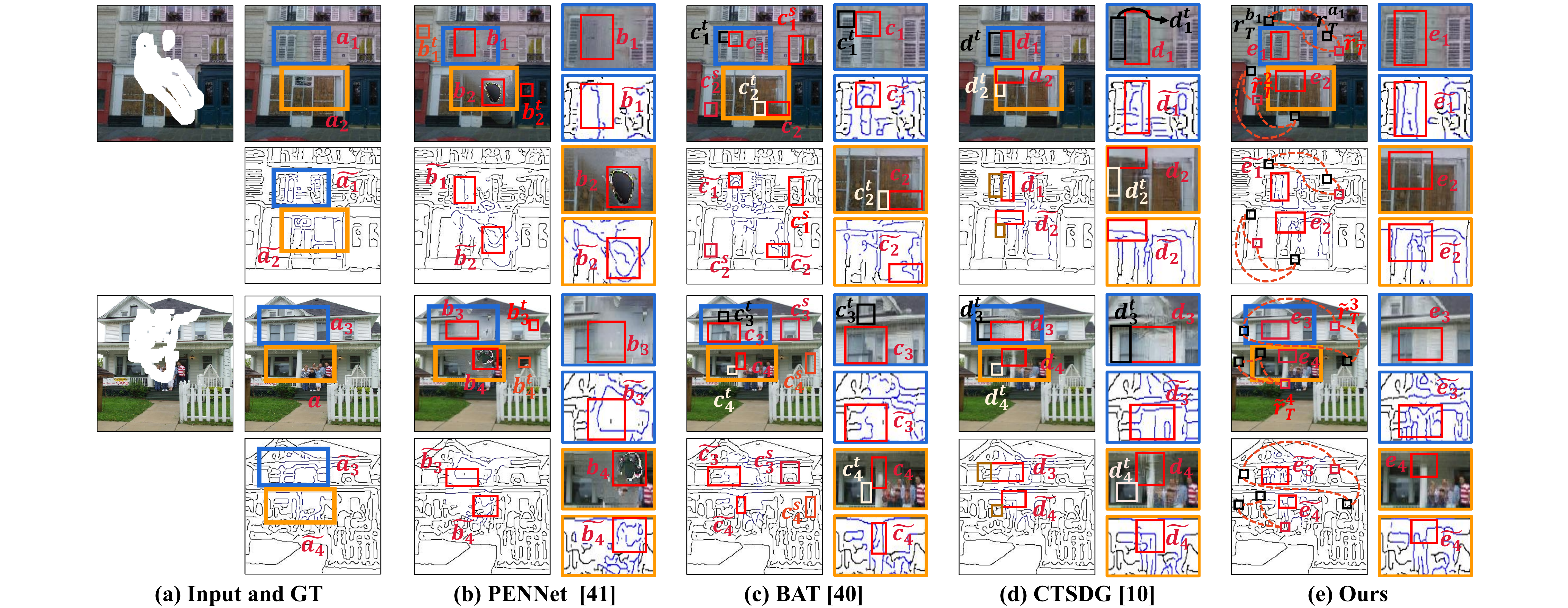}
 \caption{
 (a) Masked image and Ground Truth for texture $a_{1} \sim a_{4}$ and structure $\widetilde{a_{1}} \sim \widetilde{a_{4}}$;
  (b) The inpainting $b_{1}$ (\textit{gray artifacts}) is dominated by the \textit{local} texture of $b^{t}_{1}$ (\textit{gray wall}). Same observations for other regions;
(c) Take $c_{1}$ as example. The global structure $\widetilde{c_{1}}$ (\textit{right of window})
is well inpainted by \textit{global} structure $c^{s}_{1}$ (\textit{complete window}) among all known regions,
yet bears the \textit{local} texture $c^{t}_{1}$  (\textit{left of gray window and white frame}) ,
so $c_{1}$ (\textit{right of gray window})  suffers from \textit{white artifacts};
 (d) $\widetilde{d_{1}}$ (\textit{incomplete window}) is misguided by
 \textit{local} texture $d^{t}_{1}$ (\textit{broken window}), mistakenly inpainting $d_{1}$
 (\textit{gray window with unclear frame})  to be blurry.
 The similar problem holds for other regions; (e) Taking $e_{1}$ as example for our model, the global texture reference
  $\widetilde{r}_{T}^1$ (\textit{gray window and white frame}) is reconstructed by and encodes semantic correlations over
  $r^{a_{1}}_{T}$ (\textit{gray wall}) and $r^{b_{1}}_{T}$  (\textit{white frame and gray wall}) with large attention score
   beyond the patches with local spatial distance, to reconstruct the global structure information $\widetilde{e_{1}}$ (\textit{window and frame})
    and texture information $e_{1}$ (\textit{gray window and white frame in the gray wall}). Hence, our method achieves the better inpainting results.
}
  \label{fig:high}
\end{teaserfigure}

\maketitle

\begin{figure*}
\centering
\includegraphics[width=15.3cm]{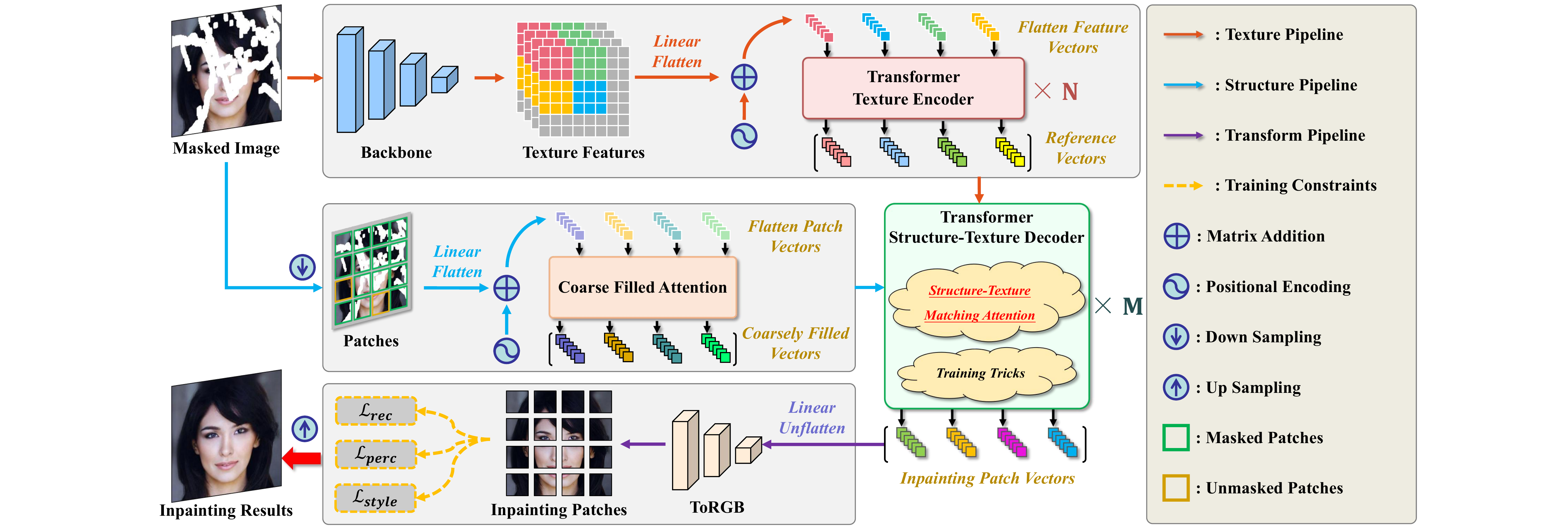}
\caption{Our transformer pipeline consists of both encoder and decoder modules. We name encoder to be Transformer Texture Encoder (TTE), as it aims to encode global texture correlations across the whole image. Coarse Filled Attention (CFA) module aims to generate the coarsely filled information over the masked regions. Transformer Structure-Texture Decoder (TSTD) exploits the global texture references set to inpaint the masked patches, where propose a novel bridge module comes up to match global texture and structure information.}
\label{fig:model}
\end{figure*}

\section{Introduction}
Image inpainting, which aims at inpainting the masked regions of the image, supports a vast range of applications, such as image editing and restoration. The pioneering diffusion-based \cite{bertalmio2000image,efros2001image} and patch-based \cite{barnes2009patchmatch, darabi2012image, hays2007scene} methods can \textit{only} be applied to inpaint masked regions with small size by simple color information from pixel level, as they failed to capture the high-level semantics for inpainting. To resolve it, substantial attention has shifted to deep models, where the convolutional neural networks (CNNs) based models \cite{pathak2016context, iizuka2017globally, liu2018image, yu2019free, Xie_2019_ICCV} follow the encoder-decoder architecture to learn high level semantic information. However, the local inductive priors for CNNs only received the filled information from the bounded known regions within the \textit{local} spatial ranges of masked regions.

To remedy such issue,  attention-based mechanisms \cite{yu2018generative, yan2019PENnet, liao2021image, zeng2021generative, wang2022progressive} have been proposed. In particular, the masked regions represented as patches are initially filled with coarse content, which serves as the query to attend to all known patches of the image, and subsequently selects the candidate with large score for substitution. Notably, \cite{yan2019PENnet} proposes a cross-layer attention module to calculate the  attention score over the feature map in deep layer, and perform the patch substitution over the low layer as per the attention score, the inpainting output is finally obtained via upsampling. Although it considers all known patches across image,  each known patches is independently considered upon masked region, such strategy will mislead the masked patch to be inpainted by only one dominant known region with the largest attention score, which may suffer from non-ideal inpainting output.

Akin to attention-based methods, transformer based models \cite{wan2021high, yu2021diverse, chen2020generative} also consider the information from all known regions. Instead of focusing on the patch pool,  it stands upon the pixel level, where each pixel within the masked regions serves as the query to attend to all other pixels from the known regions, so as to be reconstructed, which is further projected into the color vocabulary to select the most relevant color for inpainting. The inpainting pixel then joins into the known pixel pool, such process repeats until all pixels are inpainted as per a predefined order. Technically, \cite{wan2021high, yu2021diverse} propose the decoder-only transformer to capture the structure priors of pixel level and project into the visual color vocabulary for selecting the corresponding color via the dense attention module \cite{vaswani2017attention}. On one hand, it delves into all known regions rather than confirming only limited known regions for inpainting, hence it is superior to attention models. On the other hand, the pixel level cannot well capture the semantics as patch level, and therefore inferior to attention models. Besides, only the positional information is utilized to yield attention score, which is far from the texture semantic level. Furthermore, the computational burden for large number of pixels enforces transformer model to avoid quadratical complexity raised by possible self-attention module.

Going one step further to the viewpoint of texture and structure, all the above methods are basically classified as either texture-only \cite{yan2019PENnet, yu2019free} or structure-texture based methods \cite{wan2021high, yu2021diverse, zeng2021generative, xiong2019foreground, Nazeri_2019_ICCV, Li_2019_ICCV, ren2019structureflow, liao2020guidance, Liu2019MEDFE, yang2020learning, Guo_2021_ICCV,wang2021image, wang2021survey}. As the texture-only methods, such as CNNs-based \cite{yu2019free} and attention-based model \cite{yan2019PENnet} that heavily rely on the known texture information to restore the masked regions, but ignoring the structure may prohibit the reasonable texture to be recovered; worse still, the texture information exploited for inpainting only comes from the bounded known regions rather than the whole image, hence the semantic correlations among  texture from the global image may not be well captured. Different from that, the structure-texture based methods aim to generate better texture semantics for the masked regions guided by the structure constraints. Following that, the texture is recovered via varied upsampling network. To sum up, their core problem lies in how to fill the structure information into masked regions.

\begin{figure}
\centering
\includegraphics[width=7.5cm]{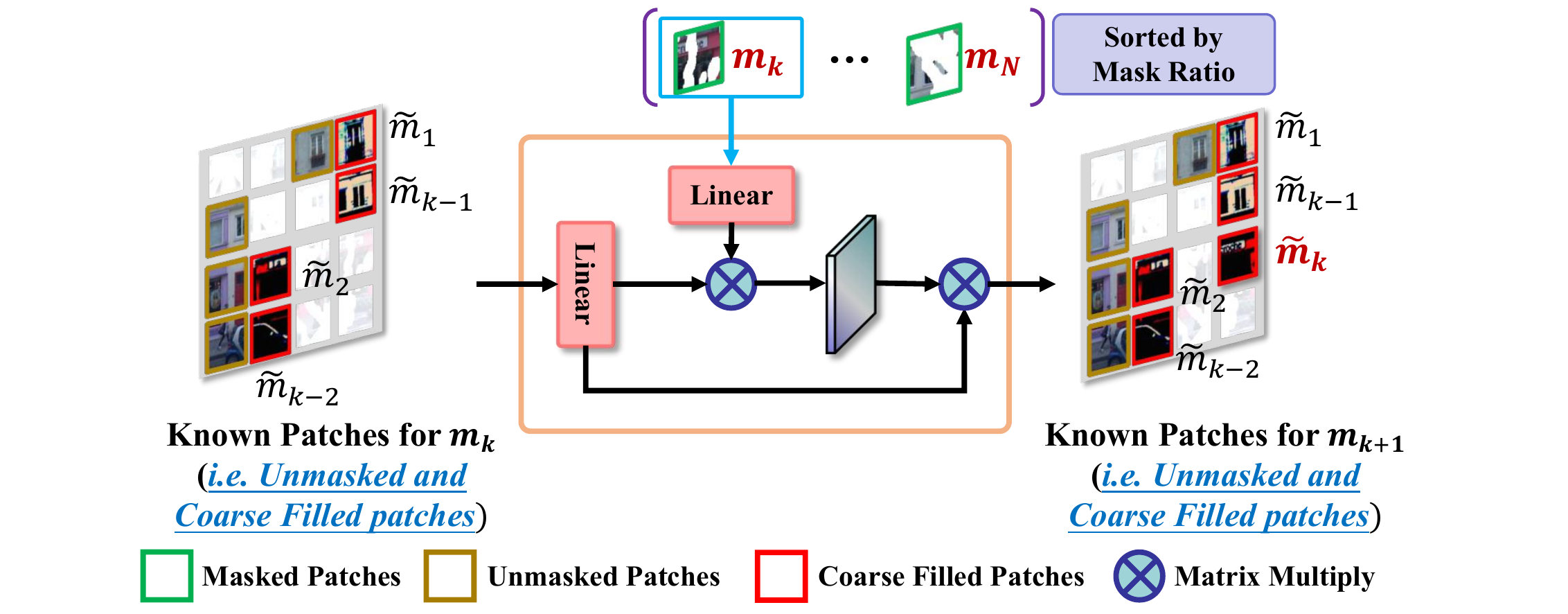}
\caption{Intuition of the Coarse Filled Attention module.}
\label{fig:cf}
\end{figure}
EdgeConnect \cite{Nazeri_2019_ICCV} restores the edge information as structure information based on edge map and masked grayscale via CNNs. The repaired edge map combines with the ground-truth masked image containing texture information to restore the masked regions via encoder-decoder model.  Wang \textit{et al.} \cite{wang2021image} adopts the CNNs model to reconstruct the masked regions of monochrome images as structure constraint, upon which, the color information as texture flow is propagated within the image via multi-scale learning. \cite{Liu2019MEDFE} follows an encoder-decoder architecture, where the encoder aims to equalize the structure feature from deep layers and texture feature from shallow layers via channel and spatial equalization, then combined as input feed to decoder to generate inpainting image. Despite the intuition of capturing the structure information, it failed to exploit the information from all known patches, we hence mildly refer it as "\textit{pseudo global structure}", and possibly misguided the non-ideal texture recovery compared with the transformer model  \cite{wan2021high, yu2021diverse}.
\cite{Guo_2021_ICCV} recently  proposed that structure and texture can mutually guide each other via a two-stream architecture based on U-Net variant.
However, it suffers from utilizing the\textit{ local } texture to guide the \textit{global} structure, leading to the blurry artifacts as illustrated in Fig.\ref{fig:high}(d).  Based on the above, it is quite beneficial as \textit{how to yield global texture and structure information} to well exploit the semantics across the whole image, and\textit{ how to match these two types of global information} for image inpainting.

As motivated above, we propose to delve globally into texture and structure for image inpainting.
To well capture the global semantic correlations from the texture for all patches across the whole image,  we adopt the transformer pipeline paired with both encoder and decoder, unlike the decoder-only transformer, the encoder encodes the correlations of the texture information regarding all the patches throughout the self-attention module, with each patch summarized as an entry over the feature map by CNNs, so as to characterize semantics. By this means, the texture information of each entry, represented as the texture reference vector, named texture reference for short, serves as the query to attend to all other texture references for reconstruction. In other words, each texture references encode varied attention scores for semantic correlation degree with all the others across the global image, yielding to a \textit{global} texture reference, yet a \textit{local} structure information caused by CNNs.
The transformer decoder aims to inpaint all the masked patches by all texture references. To this end, a Coarse Filled Attention module is developed to initially fill in all the masked patches by exploiting all the known patches. Rather than its inaccurate coarse texture information, we prefer its \textit{global} structure information benefited from exploiting all known patches of the image.  Together with all the texture references enjoying global semantic correlations, we propose a novel structure-texture matching attention module bridged by all the known patches, where the each masked patch encoding structure information serves as the query to attend to all the known patches, while each known patch serves as the query to attend to all texture references. By such means, the best matching of  these two worlds can be exploited with such transition manner, together with the adaptive patch vocabulary consisting of restored patches, to \textit{progressively} inpaint all masked patches via a \textit{probabilistic diffusion process}.  The overall pipeline is shown in Fig.\ref{fig:model}.

\begin{figure}
  \centering
  \includegraphics[width=6.5cm]{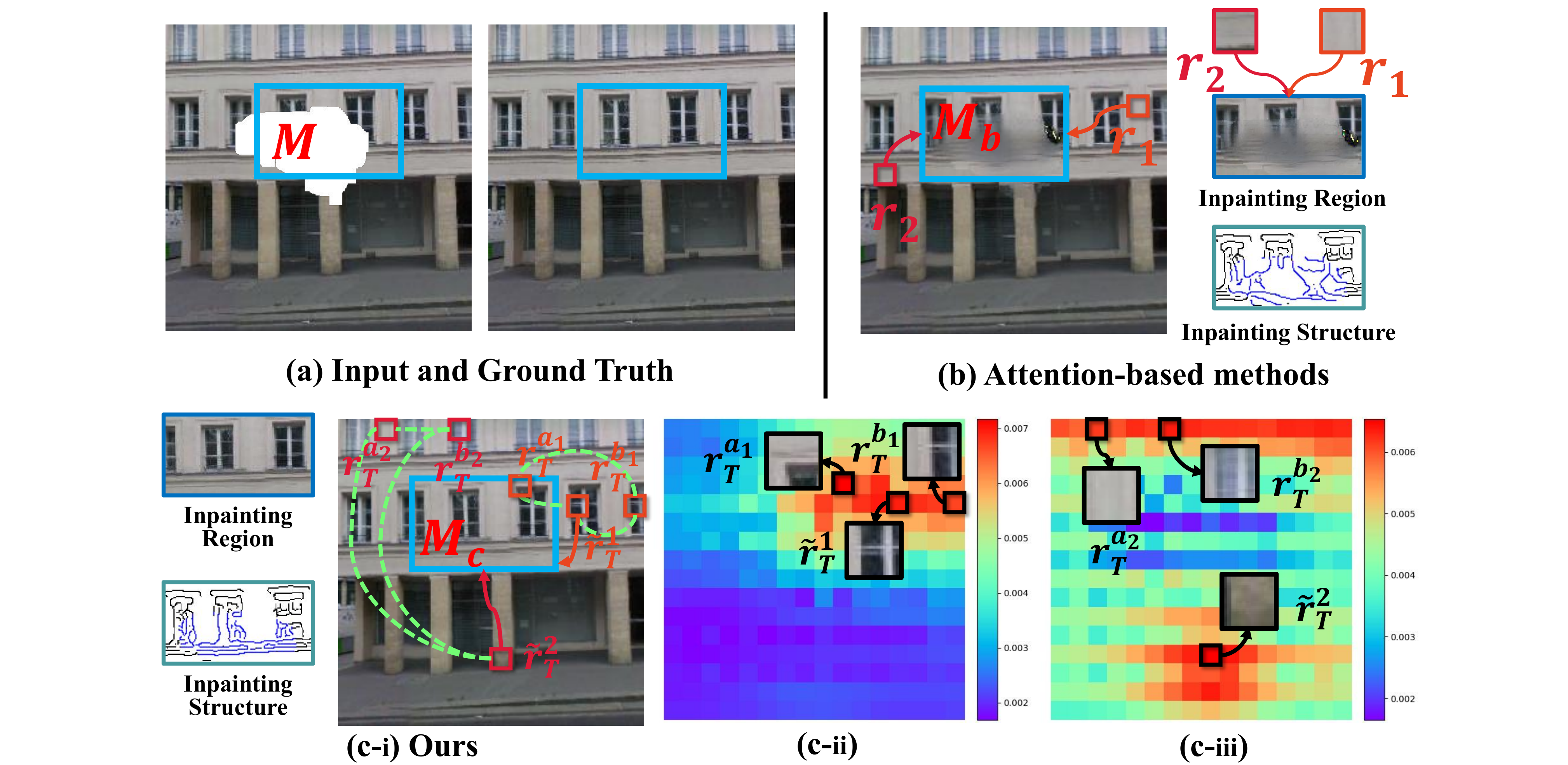}
  \caption{(a) are the input masked image and ground truth; (b) attention methods independently calculate attention score between coarse information for $M_b$ and all known regions,  to be restored by only one dominant known region;  (c-i) ours constructs global semantic correlations via texture references. (c-ii) and (c-iii) exhibit the attention map of texture references over $M_c$.
  }
	\label{fig:attention}
\end{figure}
%
\begin{figure*}
\centering
\includegraphics[width=16.5cm]{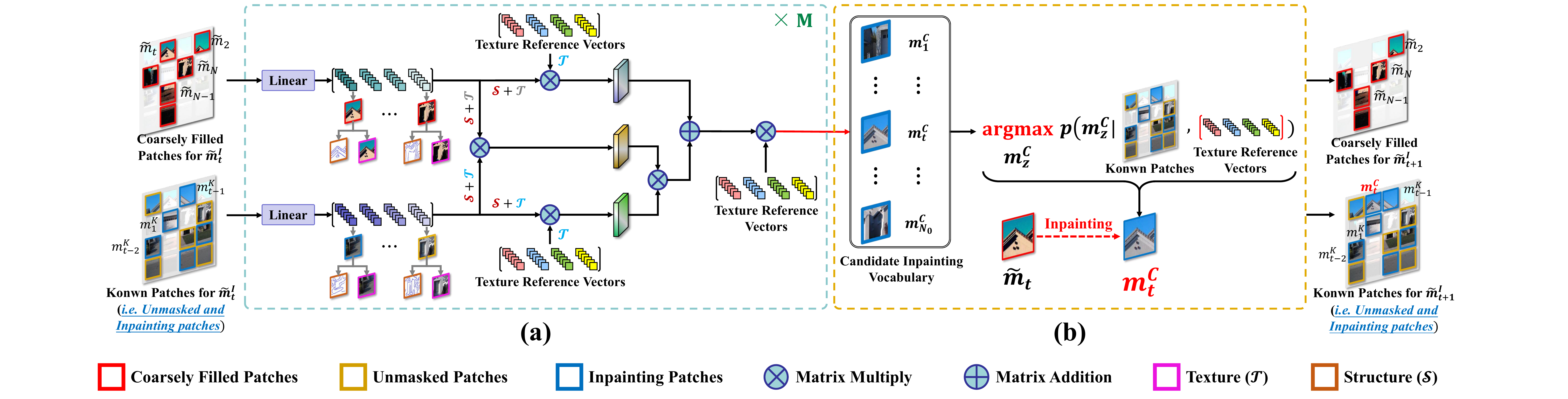}
\caption{Overall of the Transformer structure-texture Decoder. (a) The \textit{Structure-Texture Matching Attention} outputs the attention score map between each coarsely filled patch $\widetilde{m}_{t}$ and whole texture references $\widetilde{R}_{T}$, where they cannot directly match up well due to coarsely filled information. So we propose the bridge attention module over $\widetilde{R}_{T}$ based on the known patches with precise global texture and structure information, via an attention transition manner, to reconstruct $\widetilde{m}_{t}$; (b) The candidate inpainting patches over coarsely filled patches form an adaptive patch vocabulary to progressively select the most relevant candidate with the largest value of Eq.\ref{eq:Decoder3}, to expand the unmasked patch set via a probabilistic diffusion process.}
	\label{fig:decoder2}
\end{figure*}
In summary, our contributions are summarized below:

\begin{itemize}[topsep = 5pt, leftmargin = 15pt]
	\item We propose a transformer pipeline paired with both encoder and decoder, where the encoder module aims at capturing the semantic correlations of the whole images within texture references, leading to a \textit{global} texture reference set; we design a coarse filled attention module to exploit all the known image patches to fill in the masked regions, yielding a \textit{global} structure information.
	\item  To endow the decoder with the capacity of marring the best of the two worlds, \textit{i.e.,} global texture reference and structure information. we equip the decoder with a structure-texture matching attention module via an intuitive attention transition manner, where  an adaptive patch vocabulary is dynamically established for the filled patches over the masked regions via a probabilistic diffusion process.
	\item To ease the computational burden, we disclose several training tricks to overcome memory overhead for GPUs while achieve the state-of-the-art performance on typical benchmarks.
\end{itemize}

\section{Overall Framework}
Image inpainting aims to transform the input image $I_{gt} \in \mathbb{R}^{3\times H \times W}$ with mask $M$ sharing the same size, where each entry values either 1 or 0, and performing element wise multiplication over $I_{gt}$ to yield a masked image, \textit{i.e.}, $I_{m} = I_{gt} \odot M$ into a complete image $I_{out}$. In this section, we technically explain our pipeline shown in Fig.\ref{fig:model} with more details. Sec. \ref{Encoder} discusses how to encode the global semantic correlations over $I_{m}$. Based on that, Sec. \ref{decoder} discusses how to fill into the masked patches for image inpainting. We finally offer the overall loss functions in Sec. \ref{loss}.

\subsection{Transformer Texture Encoder}\label{Encoder}
To capture the textural semantic correlations across the whole image, we need to learn the explicit texture representation for each patch.
In particular, we learn that via the typical CNNs backbone, \textit{i.e.,} ResNet50 \cite{he2016deep}, to yield the high level semantic feature map $f \in \mathbb{R}^{C \times H_{0} \times W_{0}}$, where each entry of the feature map encodes the texture information for one specific patch of $I_{m}$.  It is apparent that each entry of the feature map with large size, \textit{e.g.,} $32\times 32$ w.r.t. shallow layer fails to capture high-level semantics; when going too deep, \textit{e.g.,} $8 \times 8$, each entry of the feature map bears too much semantics, resulting the texture information from one patch into mixed texture with other patches. To balance,  we set: $H_{0}=\frac{H}{16},W_{0}=\frac{W}{16}$ as \cite{carion2020end}, where we confirm the dimension for each entry as per the channel number regarding $C$, \textit{i.e.,} 2048 for ResNet50. Then we project each entry into a low dimension vector representation $r_{T}$ with dimension $d_{E} = 256$, which is performed by merging all 2048 channels first, followed by 256 filters with $1 \times 1$, to then reshape the size of feature map to be $\mathbb{R}^{d_{E} \times H_{0} \times W_{0}}$.  To preserve spatial order information, a positional embedding $E_{p_{e}} \in \mathbb{R}^{d_{E}  \times  H_{0}W_{0}}$, similar as \cite{parmar2018image},  regarding all entries are added to form the final representation $E_{T}$ for encoder.

Now we are ready to calculate the texture correlations across global image by performing the self-attention over $E_{T}$. Following transformer encoder architecture, including $N$ layers paired each layer with a multi-head self-attention (MSA) and feed forward network (FFN). For the \textit{l}th layer, it yields:
\begin{equation}
\begin{aligned}
        & H_{T}^{l} = {\rm LN}\left({\rm MSA}\left(E_{T}^{l}\right)\right)+E_{T}^{l}\\
	& E_{T}^{l+1} =  {\rm LN}\left({\rm FFN}\left(H_{T}^{l}\right)\right)+H_{T}^{l},
\end{aligned}
    \label{eq:encoder}
\end{equation}
where LN($\cdot$) denotes layer normalization and FFN($\cdot$) consists of two fully connected (FC) layers with each comprised by two
sub-layers. MSA($\cdot$) reconstructs each $r_{T}$ via multi-head self-attention module over all the others to capture the global semantic correlations, the two fully connected layer subsequently converting it to be the input of the (\textit{l}+1)th layer till the last layer, here residual connections around each of the sub-layers is employed. We formulate the Multi-head self-attention, MSA($\cdot$), over $l$th layer is computed as
\begin{equation}
\begin{aligned}
        & head^l_{j} = {\rm softmax}\left(\frac{{\rm W}^{j_{l}}_{Q}E_{T}^{l}\left({\rm W}^{j_{l}}_{K}E_{T}^{l}\right)^T}{\sqrt{d_{l}}}\right){\rm W}^{j_{l}}_{V}E_{T}^{l}\\
	& {\rm MSA} = \left[head^l_{1};\cdots;head^l_{h}\right]{\rm W}^l,
\end{aligned}
    \label{eq:encoder}
\end{equation}
where $h$ is the number of head, $d_{l}$ is the embedding dimension, ${\rm W}^{j_{l}}_{Q}$, ${\rm W}^{j_{l}}_{K}$ and ${\rm W}^{j_{l}}_{V}$ are three learnable linear projection matrices,  $1 \leq j \leq h$. ${\rm W}^l$  represents a learnable FC layer to fuse the outputs from different heads.
After the encoder layers, we can reconstruct each texture feature vector $r_{T}$ to be the reference vector $\widetilde{r}_{T}$, with all forming the reference set $\widetilde{R}_{T}$. It is easily seen that each $\widetilde{r}_{T}$  encodes the global texture correlations among all other, where the texture correlations varied across different locations.

\subsection{Transformer Structure-Texture Decoder}\label{decoder}
Besides texture reference set $\widetilde{R}_{T}$ from encoder, it is crucial to model the representation regarding the masked patches.
Unlike the existing decoder-only transformer over pixel level, we need to consider the patch size for the semantics matching with $\widetilde{R}_{T}$.
We simply reshape $I_{m}$ to $I_{m}^{\prime}$ to get low resolution image such that $I_{m}^{\prime} \in \mathbb{R}^{3\times \frac{H}{4} \times \frac{W}{4}}$ to strength the global structure information, while achieving a medium patch size. Inspired by \cite{dosovitskiy2020image, lee2021vitgan}, we convert $I_{m}^{\prime}$ to a sequence of flattened 2D patches $I_{p}^{\prime} \in \mathbb{R}^{N_{0} \times \left(3 \times P \times P\right)}$, with $P \times P$ resolution for each patch, and $N_{0}$ is the patch number of patches, then we flatten the patches and map to $d_{D}$ dimensions with a trainable linear matrix as the patch embedding. For both known patches and unknown patches, the extra spatial position embedding $E_{p_{d}}\in\mathbb{R}^{d_{D}  \times  N_{0}}$ is added into the flattened patches for spatial order preservation.

Before discussing the correlations between $\widetilde{R}_{T}$ and masked regions, we need to fill the masked regions with coarse information based on the known regions.  Unlike \cite{yu2018generative} relying on CNNs with local inductive prior to exploit the known patches to fill in the coarse content , we propose a globally filled attention mechanism to exploit all the known regions of the image. To ease the understanding, we illustrate it in Fig.\ref{fig:cf}. To coarsely fill in the $k^{th}$ masked patch $m_{k}$, we leverage the known patch and previous \textit{k}-1 coarsely filled patches. Specifically, all masked patches are sorted by ascending order by the mask ratio to be filled with coarse content, and reconstruct each of them throughout the attention over the known patches and other coarsely filled patches $\widetilde{m}_{i}, 1 \leq i \leq k-1$, both of them form the set $P_{k-1}$, together with the filled $\widetilde{m}_{k}$ to further become $P_{k}$. The coarsely filled $\widetilde{m}_{k}$   w.r.t. ${m}_ {k}$  is reconstructed via attention score over $P_{k-1}$ as:
\begin{equation}
\begin{aligned}
        & \widetilde{m}_{k} = {\rm softmax}\left(\frac{{\rm W}^C_{Q}m_{k}\left({\rm W}^C_{K}P_{k-1}\right)^T}{\sqrt{d_{m}}}\right){\rm W}^C_{V}P_{k-1}\\
		& P_{k} = P_{k-1} \cup \widetilde{m}_{k},
\end{aligned}
    \label{eq:encoder}
\end{equation}
where $d_{m}$ is the embedding dimension, ${\rm W}^C_{Q}$, ${\rm W}^C_{K}$ and ${\rm W}^C_{V}$ are three learnable linear projection matrices.
\begin{figure}
  \centering
  \includegraphics[width=7cm]{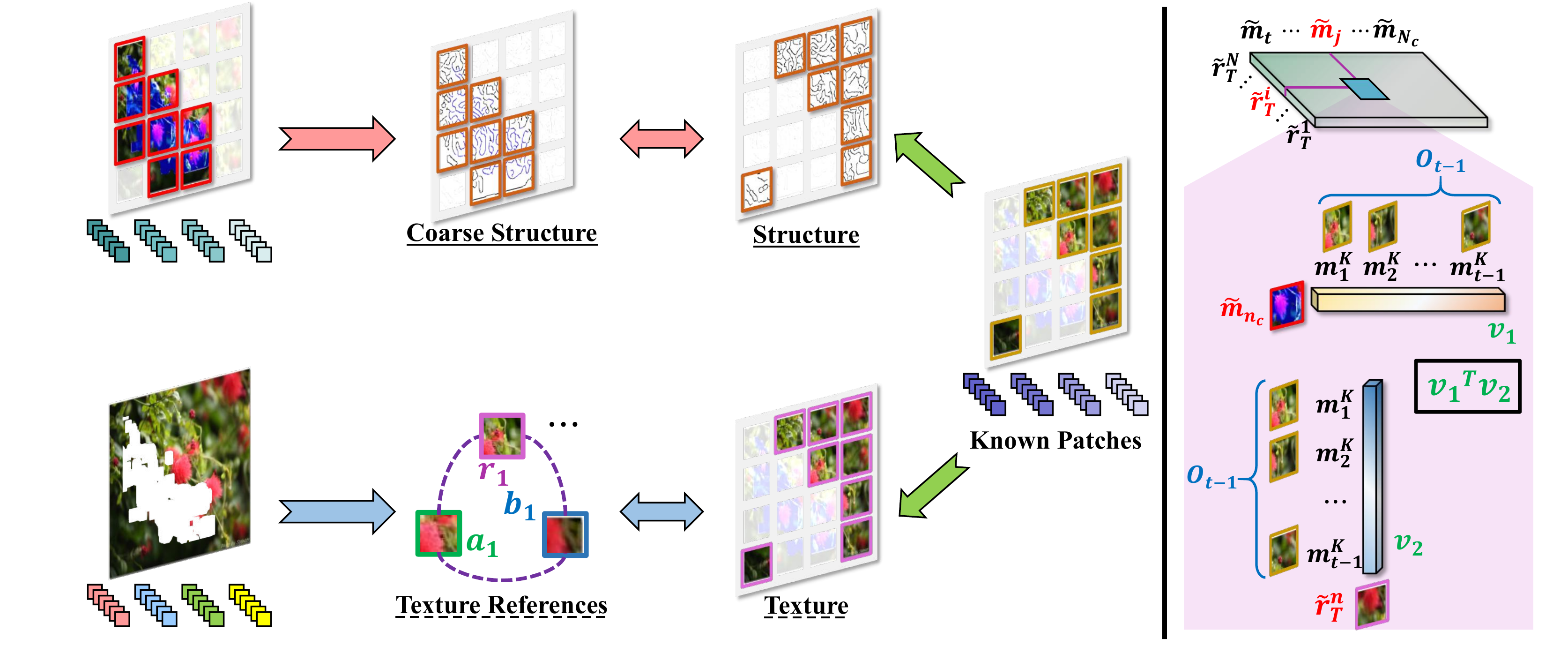}
  \caption{Intuition of the bridge module. The masked patches filled with coarse texture  but fine global structure information serve as the query to attend to all known patches, while each known patch servers as the query to attend to all texture references via an attention transition manner.}
	\label{fig:bridge}
\end{figure}

Now we discuss how to inpaint each $\widetilde{m}_{t}$  by selecting the desirable $\widetilde{r}_{T}$ from $\widetilde{R}_{T}$. As previously discussed,  the conventional attention mechanism shown in Fig.\ref{fig:attention}(b) independently calculates the attention score between masked and each individual known regions, misleading the masked patch to be inpainted by only one dominant known region. Unlike it, our proposed method as shown in Fig.\ref{fig:attention}(c) exploits the global texture semantic correlations over $\widetilde{R}_{T}$ for inpainting. We observe that both $\widetilde{r}_{T}$ and $\widetilde{m}_{t}$ are formed by exploiting unmasked regions across the whole image.
It is apparent that $\widetilde{r}_{T}$ well captures \textit{global texture} information, which is much more precise than the coarsely filled texture information for $\widetilde{m}_{k}$. However, the structure information by exploiting all the unmasked patches for each $\widetilde{m}_{t}$ is much better, and strengthened via downsampling. Motivated by this, beyond reconstructing $\widetilde{m}_{t}$ via attention module directly from $\widetilde{R}_{T}$. We also propose to serve the unmasked regions, which enjoys both the ideal texture and structure information, as a bridge module to match $\widetilde{r}_{T}$ and  $\widetilde{m}_{k}$, as depicted in Fig.\ref{fig:bridge}.

For $M$ layers of the decoder, each layer consists of two sub-layers: a structure-texture matching attention (STMA) module and a FFN function with two fully connected layers, which subsequently converts it to be the input of the (\textit{l}+1)th layer till the $M$th layer as similar as encoder.
We equally employ residual connections around each of them. For $\widetilde{m}_{t}, t \leq N_{0}$ over the $l$th layer, it yields:
\begin{equation}
\begin{aligned}
        & \widetilde{h}^{l}_{t} = {\rm LN}\left({\rm STMA}\left(\widetilde{m}^{l}_{t}\right)\right)+\widetilde{m}^{l}_{t}\\
	& \widetilde{m}^{l+1}_{t} =  {\rm LN}\left({\rm FFN}\left(\widetilde{h}^{l}_{t}\right)\right)+\widetilde{h}^{l}_{t},
\end{aligned}
    \label{eq:decoder_layer}
\end{equation}
where STMA($\cdot$) denotes the structure-texture matching attention, to obtain the attention score map between $\widetilde{m}_{t}$ and $\widetilde{R}_{T}$ based on known regions including inpainting patches $m^K_{j}, 1 \leq j \leq t-1$, denoted as $O_{t-1}$. The STMA($\cdot$) performs over the $l$th layer as:
\begin{equation}
\begin{aligned}
        & S^{d_{l}}_{t} = \left(\frac{{\rm W}^{d_{l}}_{Q}\widetilde{m}^{l}_{t}\left({\rm W}^{d_{l}}_{K}\widetilde{R}_{T}\right)^T}{\sqrt{d_{i}}}\right),\\
\end{aligned}
    \label{eq:Decoder}
\end{equation}
where $S^{d_{l}}_{t}$ directly computes the attention scores between $\widetilde{m}_{t}$ and $\widetilde{R}_{T}$. ${\rm W}^{d_{l}}_{Q}$ and ${\rm W}^{d_{l}}_{K}$ are learnable linear projection matrices and  $d_{i}$ is the embedding dimension. As aforementioned, the coarse texture information from $\widetilde{m}_{t}$ may not directly match $\widetilde{R}_{T}$ well, so we propose bridging attention module over $\widetilde{R}_{T}$ based on the unmasked patches to reconstruct  $\widetilde{m}_{t}$ over the $l$th layer below:
\begin{small}
\begin{equation}
\begin{aligned}
		&S^{b_{l}}_{t} =  \left(\frac{{\rm W}^{b_{l}}_{Q_{c}}\widetilde{m}^{l}_{t}\left({\rm W}^{b_{l}}_{K_{c}}O^{l}_{t-1}\right)^T}{\sqrt{d_{c}}}\right) \left(\frac{{\rm W}^{b_{l}}_{Q_{r}}O^{l}_{t-1}\left({\rm W}^{b_{l}}_{K_{r}}\widetilde{R}_{T}\right)^T}{\sqrt{d_{r}}}\right),\\
\end{aligned}
    \label{eq:Decoder1}
\end{equation}
\end{small}
where $S^{b_{l}}_{t}$ denotes the bridge attention module. ${\rm W}^{b_{l}}_{Q_{c}}$, ${\rm W}^{b_{l}}_{K_{c}}$, ${\rm W}^{b_{l}}_{Q_{r}}$ and ${\rm W}^{b_{l}}_{K_{r}}$ are learnable linear projection matrices,  $d_{c}$ and $d_{r}$ are the embedding dimension.

\begin{figure}
  \centering
  \includegraphics[width=7.5cm]{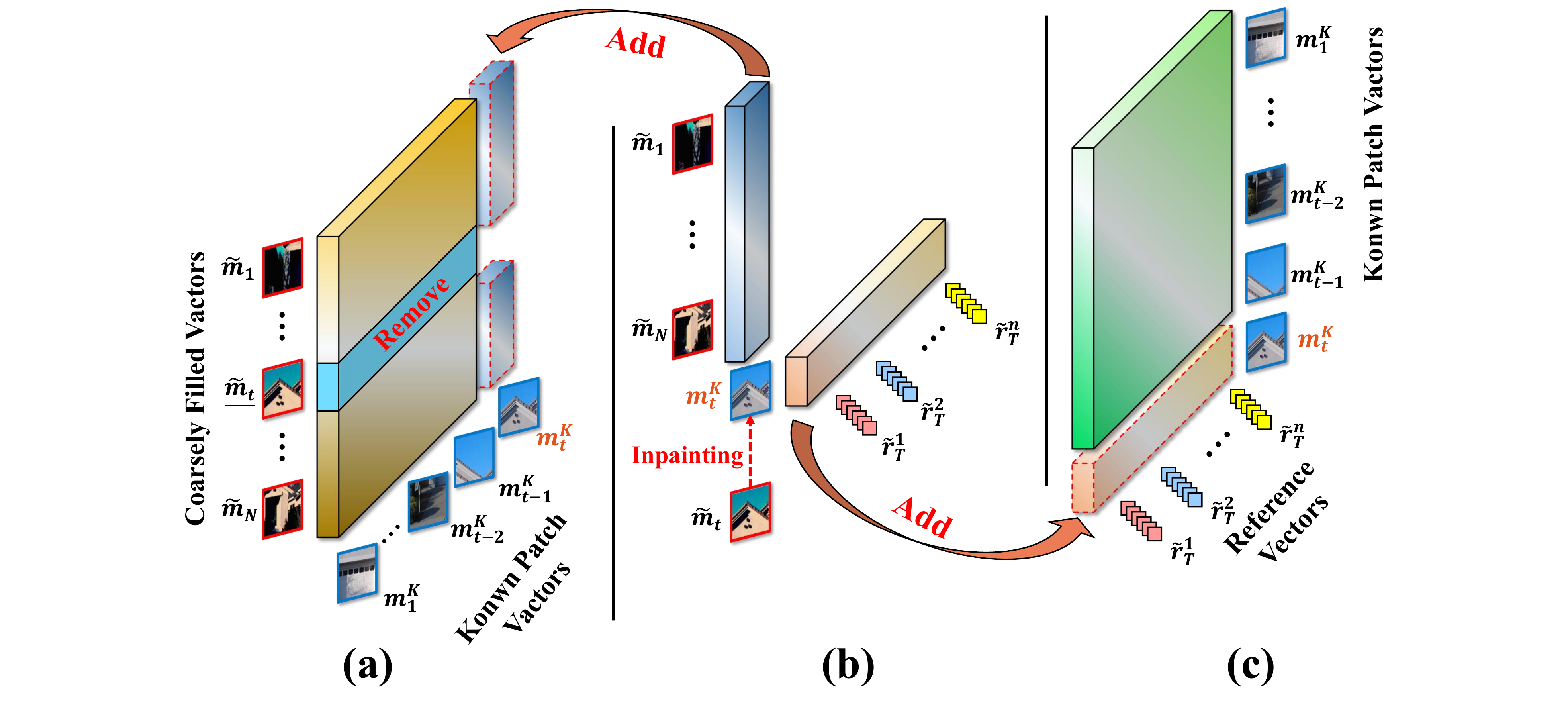}
  \caption{Intuition of the training tricks for computational efficiency. }
  \label{fig:save}
\end{figure}
We remark that Eq.\ref{eq:Decoder1} implies an \textit{attention transition manner} by serving $\widetilde{m}^{l}_{t}$ as the query to attend to $O^{l}_{t-1}$, each of $O^{l}_{t-1}$ further serves as the query to attend to $\widetilde{R}_{T}$, to finally reconstruct $\widetilde{m}^{l}_{t}$; note that we only achieve the attention score with the known patches as query to attend to $\widetilde{R}_{T}$ to reconstruct $\widetilde{m}^{l}_{t}$, rather than $O^{l}_{t-1}$ to further reconstruct $\widetilde{m}^{l}_{t}$, as it is apparent that the known patches are ideal ground truth, and should \textit{not} be reconstructed. Combining Eq.\ref{eq:Decoder} with \ref{eq:Decoder1}, it has:
\begin{equation}
\begin{aligned}
	& m^{C_{l}}_{t} = \lambda{\rm softmax}\left( S^{d_{l}}_{t}\right){\rm W}^{d_{l}}_{V}\widetilde{R}_{T}+ \left(1- \lambda\right){\rm softmax}\left(S^{b_{l}}_{t}\right){\rm W}^{b_{l}}_{V}\widetilde{R}_{T},
\end{aligned}
    \label{eq:Decoder2}
\end{equation}
where ${\rm W}^{d_{l}}_{V}$ and ${\rm W}^{b_{l}}_{V}$ are learnable linear projection matrices. Following this, we require a decoder vocabulary over the patch level, from which we generate each inpainting patch. Particularly, each coarsely filled patch $\widetilde{m}^{M}_{z}$ is reconstructed via Eq.\ref{eq:Decoder2} over the final $M$th layer to become $m^{C}_{z}$, while form vocabulary set $m^C$, from which we select the most relevant candidate $m^K_{t}$ to become the final inpainting patch with the largest probability via Eq.\ref{eq:Decoder3}.
\begin{small}
\begin{equation}
\begin{aligned}
	m^K_{t}  &= \mathop{{\rm arg\,max}}_{m^{C}_{z}} p\left(m^{C}_{z} | O_{t-1}, \widetilde{R}_{T}\right)\\
	&=  \mathop{{\rm arg\,max}}_{m_{z}^C} \frac {\lambda\left \| {\rm softmax}\left( S^{d_{M}}_{z}\right)\right \|_{1} +\left(1- \lambda\right)\left \|{\rm softmax}\left(S^{b_{M}}_{z}\right) \right \|_{1}}{ \sum\limits_{t\leq j \leq N_{c}}\left(\lambda\left \|{\rm softmax}\left( S^{d_{M}}_{j}\right)\right \|_{1} +\left(1- \lambda\right)\left \|{\rm softmax}\left(S^{b_{M}}_{j}\right) \right \|_{1}\right)},\\
		&O_{t} = O_{t-1} \cup m_{t}^K,
\end{aligned}
    \label{eq:Decoder3}
\end{equation}
\end{small}
where softmax$(S^{d_{M}}_{z})$, softmax$(S^{b_{M}}_{z})$ are the output vector representation with 256 dimensions via the last $M$th layer, the $i$th entry softmax$(S^{d_{M}}_{z})$(i) and softmax$(S^{b_{M}}_{z})$(i) denote the attention score between the $z$th coarsely filled patch $\widetilde{m}^{M}_{z}$ and the $i$th  texture reference $\widetilde{r}_{T}^{i}$ via Eq.\ref{eq:Decoder} and \ref{eq:Decoder1}, $\left \| \cdot \right \|_{1}$ sums up all entries over $\widetilde{R}_{T}$, which help reconstruct all $\widetilde{m}^{M}_{z}$ to form $m^{C}_{z}$ in $m^C$; $N_{c}$ is the number of elements for $m^C$. The most relevant candidate $m_{z}^C$ is selected to be the winner $m_t^K$ in the $t$th iteration with the largest attention score summation, via $\left \| \cdot \right \|_{1}$ norm, over $\widetilde{R}_{T}$, \textit{i.e.,} the summation of attention score over  $\widetilde{R}_{T}$ for $\widetilde{m}^{M}_{z}$, to join the known patch set to expand $O_{t-1}$ to $O_{t}$ via a \textit{probabilistic diffusion process} throughout Eq.\ref{eq:Decoder3}, while further help select the others candidate to be inpainted in the $(t+1)$th iteration, and end up with all patches to be inpainted. We remark that our decoder patch vocabulary $m^C$ is adaptively constructed based on coarsely filled patches, and dynamically updated. The overall architecture for our transformer decoder is shown in Fig.\ref{fig:decoder2}.
\\
\textbf{Computational efficiency.} One may wonder the computational complexity caused by the attention module for each iteration. We clarify it to be efficient as shown in Fig.\ref{fig:save}, the attention map among coarsely filled patch set, known patch set and texture reference set can be computed off the shelf. When restoring a coarsely filled patch into inpainting patch, there is no need to recalculate whole the attention map between different sets, but only remove the corresponding row of attention map meanwhile make up the column for the known patch w.r.t. the coarsely filled patch as per Fig.\ref{fig:save}(a) and \ref{fig:save}(b). We also supplement the new row to the attention map between newly inpainting patch and texture reference set in \ref{fig:save}(c). After inpainting most of candidate patches from vocabulary $m^{C}$, there is no need to repeat the probabilistic diffusion process via Eq.\ref{eq:Decoder3}, especially for the candidate with quite small ratio for coarse content, we simply average such content with the neighborhood patches to further reduce the attention complexity to $\widetilde{R}_{T}$ and $m^K$.

\begin{table*}
    \centering
\tiny
    \caption{Quantitative results with varied mask ratios under ${\ell_{1}}$, PSNR, SSIM, and FID on PSV \cite{doersch2012makes} , CelebA-HQ \cite{liu2015deep} and Places2 \cite{zhou2017places} with irregular mask dataset\cite{liu2018image} ($\uparrow$ : Higher is better; $\downarrow$ : Lower is better; - : no reported results from the methods). The two best scores are colored by
\textcolor{red}{red} and \textcolor{blue}{blue}.}
    \label{tab:compare}
    \begin{tabular}{@{}c|l|c|cccc|cccc|cccc|c@{}}
    \toprule
	\multicolumn{3}{c|}{Mask Ratio}  &\multicolumn{4}{c|}{20-40\%}  &\multicolumn{4}{c|}{40-60\%}  &\multicolumn{4}{c|}{Random} & 20-40\% \\\midrule

    Dataset &Method &Venue &${\ell_{1}}$(\%)$\downarrow$ &PSNR$\uparrow$ &SSIM$\uparrow$ &FID$\downarrow$ & ${\ell_{1}}$(\%)$\downarrow$ &PSNR$\uparrow$ &SSIM$\uparrow$ &FID$\downarrow$ &${\ell_{1}}$(\%)$\downarrow$ &PSNR$\uparrow$ &SSIM$\uparrow$ &FID$\downarrow$ & User Study\\ \midrule
	\midrule
		\multirow{10}*{{Places2 \cite{zhou2017places}}}
	&EC \cite{Nazeri_2019_ICCV} &ICCV' 19 &2.20 &26.52 &0.880 &25.64  &4.38 &22.23 &\textcolor{blue}{0.731} &39.27 &2.93 &25.51 &0.831 &30.13 &2\\
	&GC \cite{yu2019free} &ICCV' 19 &\textcolor{blue}{2.15} &\textcolor{blue}{26.53} &\textcolor{blue}{0.881} &24.76   &4.40 &21.19 &0.729 &39.02   &2.80 &25.69 &0.834 &29.98  &11\\
	&PIC \cite{zheng2019pluralistic}&CVPR' 19 &2.36 &26.10 &0.865 &26.39   &5.07 &21.50 &0.680 &49.09   &3.15 &25.04 &0.806 &33.47  &-\\
	&HiFill \cite{yi2020contextual} &CVPR' 20 &- &- &- &-   &- &- &- &-   &- &24.35 &0.867 &-&-  \\
	&MEDFE \cite{Liu2019MEDFE} &ECCV' 20 &2.24 &26.47 &0.877 &26.98   &4.57 &\textcolor{blue}{22.27} &0.717 &45.46   &2.91 &25.63 &0.827 &31.40 &5 \\
	&CTSDG \cite{Guo_2021_ICCV}&ICCV' 21 &- &25.97 &0.759 &-   &- &22.23 &0.561 &-   &- &- &- &- &19 \\
	&ICT \cite{wan2021high} &ICCV' 21 &2.44 &26.50  &0.880  &21.60    &\textcolor{blue}{4.31} &22.22 &0.724 &33.85   &\textcolor{blue}{2.67} &\textcolor{blue}{25.79} &0.832 &25.42  &\textcolor{blue}{31}\\
	&EII \cite{wang2021image}&CVPR' 21 &- &- &- &-   &- &- &- &-   &- &24.58 &\textcolor{blue}{0.880} &- &- \\
	&BAT \cite{yu2021diverse}&ACM MM' 21 &\textcolor{blue}{2.15} &26.47 &0.879 &\textcolor{blue}{17.78}   &4.64 &21.74 &0.704 &\textcolor{blue}{32.55}   &2.84 &25.69 &0.826 &\textcolor{blue}{22.16} &25 \\
	&Ours &- &\textcolor{red}{1.85} &\textcolor{red}{27.77}&\textcolor{red}{0.946} &\textcolor{red}{14.75}  &\textcolor{red}{4.00}&\textcolor{red}{22.99}&\textcolor{red}{0.858}& \textcolor{red}{30.66}  &\textcolor{red}{2.40}&\textcolor{red}{27.93}&\textcolor{red}{0.910} &\textcolor{red}{18.84} &\textcolor{red}{107}\\
\midrule
	\multirow{7}*{{CelebA \cite{liu2015deep}}}
	&EC \cite{Nazeri_2019_ICCV}&ICCV' 19 &2.19 &26.60 &0.923 &9.06   &4.71 &22.14 &0.823 &16.45   &3.40 &24.45 &0.877 &12.46 &7 \\
	 &GC \cite{yu2019free}&ICCV' 19  &2.70 &25.17 &0.907 &14.12   &5.19 &21.21 &0.805 &22.80   &3.88 &23.32 &0.858 &18.10 &16 \\
	 &PIC \cite{zheng2019pluralistic}&CVPR' 19 &2.50 &25.92 &0.919 &10.21   &5.65 &20.82 &0.780 &18.92   &4.00 &23.46 &0.852 &14.12 &- \\
	 &HiFill \cite{yi2020contextual}&CVPR' 20 &- &- &- &-   &- &- &- &-   &- &27.20 &0.936 &- &-  \\
	 &EII \cite{wang2021image}&CVPR' 21 &- &- &- &-   &- &- &- &-   &- &\textcolor{blue}{27.51} &\textcolor{blue}{0.945} &-&-  \\
	 &BAT \cite{yu2021diverse}&ACM MM' 21 &\textcolor{blue}{1.91} &\textcolor{blue}{27.82} &\textcolor{blue}{0.944} &\textcolor{blue}{6.32}   &\textcolor{blue}{4.57} &\textcolor{blue}{22.40} &\textcolor{blue}{0.834} &\textcolor{blue}{12.50}   &\textcolor{blue}{3.18} &25.21 &0.890 &\textcolor{blue}{9.33} &\textcolor{blue}{44} \\
	&Ours &-   &\textcolor{red}{{1.72}}&\textcolor{red}{{28.83}} &\textcolor{red}{{0.961}} &\textcolor{red}{{5.74}}    &\textcolor{red}{{4.16}} &\textcolor{red}{{23.44}} &\textcolor{red}{{0.877}} &\textcolor{red}{{11.37}}   &\textcolor{red}{{2.70}}  &\textcolor{red}{{29.13}} &\textcolor{red}{{0.949}} &\textcolor{red}{{8.80}} &\textcolor{red}{133} \\

	\midrule
	\multirow{10}*{{PSV \cite{doersch2012makes}}}
	&EC \cite{Nazeri_2019_ICCV}&ICCV' 19 &2.63 &26.76 &0.874 &42.81    &\textcolor{blue}{5.18} &22.77 &\textcolor{blue}{0.712} &72.78    &3.63 &25.04 &0.806 &55.29 &4\\
 	&GC \cite{yu2019free}&ICCV' 19 &3.56 &23.95 &0.796 &71.02    &6.31 &20.83 &0.631 &98.32    &4.64 &22.61 &0.727 &84.16 &14\\
	&PIC \cite{zheng2019pluralistic}&CVPR' 19 &3.43 &24.80 &0.817 &56.83    &7.47 &20.12 &0.57 &90.91    &4.94 &22.97 &0.718 &72.16&-\\
	&HiFill \cite{yi2020contextual}&CVPR' 20 &- &- &- &-    &- &- &- &-    &- &26.24 &0.866 &-&-\\
	&MEDFE \cite{Liu2019MEDFE}&ECCV' 20 &\textcolor{blue}{2.29} &\textcolor{blue}{27.64} &\textcolor{blue}{0.898} &36.84    &5.54 &22.01 &0.704 &77.26    &\textcolor{blue}{3.58} &25.24 &0.818 &54.99 &9\\
 	&CTSDG \cite{Guo_2021_ICCV}&ICCV' 21 &- &27.48 &0.777 &-    &- &\textcolor{blue}{22.89} &0.573 &-    &- &- &- &- &21\\
 	&EII \cite{wang2021image}&CVPR' 21 &- &- &- &-    &- &- &- &-    &- &\textcolor{blue}{26.75} &\textcolor{blue}{0.868} &-&-\\
 	&BAT \cite{yu2021diverse}&ACM MM' 21 &2.70 &26.52 &0.864 &\textcolor{blue}{36.19}    &5.83 &21.89 &0.678 &\textcolor{blue}{64.2}    &3.96 &24.50 &0.786 &\textcolor{blue}{48.19} &\textcolor{blue}{34}\\
	&Ours &-  &\textcolor{red}{{2.04}} &\textcolor{red}{{28.00}}  &\textcolor{red}{{0.929}} &\textcolor{red}{{25.21}}    &\textcolor{red}{{4.64}} &\textcolor{red}{{22.91}} &\textcolor{red}{{0.805}} &\textcolor{red}{{57.9}}    &\textcolor{red}{{2.79}}  &\textcolor{red}{{28.90}} &\textcolor{red}{{0.874}} &\textcolor{red}{{36.69}} &\textcolor{red}{118}\\
\bottomrule
    \end{tabular}
  \label{table:compare}
\end{table*}

\subsection{Loss Functions}\label{loss}
After all masked patches are inpainted, we obtain the reconstructed vector set with each locating in 384 dimensional feature space, which are further restored to RGB image following \cite{jiang2021transgan}, denoted by  $I_{out}^{\prime} \in \mathbb{R}^{3\times \frac{H}{4} \times \frac{W}{4}}$. We basically follow \cite{Liu2019MEDFE, Nazeri_2019_ICCV} over typical losses to measure reconstruction error between inpainting images $I_{out}^{\prime}$ and downsampling ground truth $I_{gt}^{\prime}$, such as reconstruction loss $\mathcal{L}_{rec}$, perceptual loss $\mathcal{L}_{prec}$, style loss $\mathcal{L}_{style}$ and adversarial loss $\mathcal{L}_{adv}$.  Afterwards, we perform the upsampling over the $I_{out}^{\prime}$ to be $I_{out} \in \mathbb{R}^{3\times H \times W}$ through a adversarial network \cite{yu2021diverse} guided by $\mathcal{L}_{adv}$,  which will not added into our final loss function. Specifically, we aim to learn desirable $I_{out}^{\prime}$ via the followings: \\
\textbf{Reconstruction Loss.} We employ the ${\ell_{1}}$ loss to measure the pixel wise difference between the ground truth $I_{gt}^{\prime}$ after downsampling and our inpainting output $I_{out}^{\prime}$, then yields the following:
\begin{equation}
\begin{aligned}
	& \mathcal{L}_{rec} = \left \| I_{out}^{\prime} - I_{gt}^{\prime} \right \|_{1}.
\end{aligned}
    \label{eq:loss1}
\end{equation}
\textbf{Perceptual Loss.} To simulate human perception of images quality, we utilize the perceptual loss by defining a distance measure between activation feature maps of a pre-trained network over our inpainting output and ground truth, we have the following:
\begin{equation}
\begin{aligned}
	& \mathcal{L}_{prec} = \mathbb{E}\left[ \sum\limits_{i}\frac{1}{N_{i}}\left \| \phi_{i}\left(I_{out}^{\prime}\right) - \phi_{i}\left(I_{gt}^{\prime}\right) \right \|_{1}\right],
\end{aligned}
    \label{eq:loss2}
\end{equation}
where $\phi_{i}$ is the activation feature map $N_i$ with the size of $C_{i} \times H_{i} \times W_{i}$ of the $i$th layer of VGG backbone. In our work, $\phi_{i}$ represents the activation maps from layers ReLu1\_1, ReLu2\_1, ReLu3\_1, ReLu4\_1 and ReLu5\_1.\\
\textbf{Style Loss.} The above activation maps from Eq.\ref{eq:loss2} are further utilized to compute style loss to measure the differences between covariances of the activation maps to mitigate  "checkerboard" artifacts. Given the $j$th layer activation feature map of VGG backbone , the style loss is formulated below:
\begin{equation}
\begin{aligned}
	& \mathcal{L}_{style} = \mathbb{E}_{j}\left[ \left\| G_{j}^\phi\left(I_{out}^{\prime}\right) - G_{j}^\phi\left(I_{gt}^{\prime}\right) \right\|_{1}\right],
\end{aligned}
    \label{eq:loss3}
\end{equation}
where $G_{j}^\phi \in R^{C_{j} \times C_{j}}$ is a Gram matrix as the selected activation maps.\\
\textbf{Overall Loss.} Based on the above, we finally offer the overall loss as Eq.\ref{eq:loss4} that is minimized to train the overall transformer pipeline:
\begin{equation}
\begin{aligned}
	& \mathcal{L}_{tran} = \lambda_{r}\mathcal{L}_{rec} + \lambda_{p}\mathcal{L}_{prec} +\lambda_{s}\mathcal{L}_{style}.
\end{aligned}
    \label{eq:loss4}
\end{equation}
For our experiments, we set $\lambda_{r} = 10$; following \cite{Liu2019MEDFE}, we set $\lambda_{p}=0.1$ and $\lambda_{s}=250$. Finally, the upsampling operation is performed over the $I_{out}^{\prime}$ to be final inpainting out $I_{out}$.

\begin{table}
    \centering
	\tiny
    \caption{Ablation study over PSV (best results are boldface).}
    \label{tab:compare}
    \begin{tabular}{@{}c|ccc|ccc@{}}
    \toprule
	 Models  &TTE&Eq.\ref{eq:Decoder}&Eq.\ref{eq:Decoder1}& ${\ell_{1}}$(\%)$\downarrow$ &PSNR$\uparrow$ &SSIM$\uparrow$ \\
	\midrule
	\midrule
	 Baseline &-&\checkmark&-&6.79 &21.99 &0.887 \\
	Baseline + Transformer Texture Encoder&\checkmark&\checkmark&- & 6.51 &22.59 &0.904 \\
	Baseline + Bridge Module  &-&\checkmark&\checkmark&6.61 &22.68 &0.907 \\
	Ours   &\checkmark&\checkmark&\checkmark&\textbf{6.44} &\textbf{22.83} &\textbf{0.915} \\
\bottomrule
    \end{tabular}
  \label{table:ab}
\end{table}
\begin{figure}
  \centering
  \includegraphics[width=6cm]{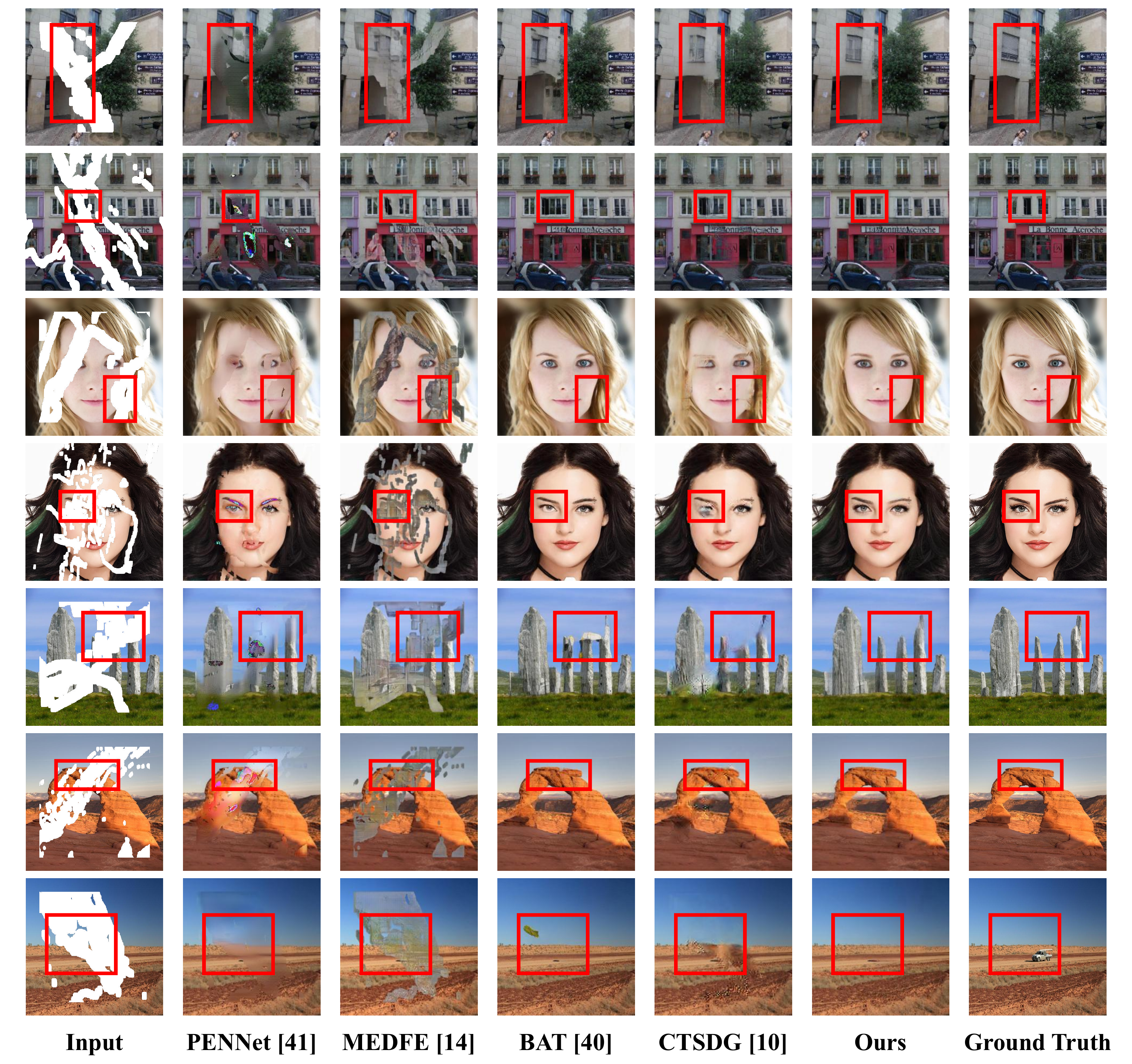}
  \caption{Visual comparison between our method and the competitors (zoom-in to see more details). Our method generates more realistic image inpainting over PSV \cite{doersch2012makes} , CelebA-HQ \cite{liu2015deep} and Places2 \cite{zhou2017places} with irregular masks \cite{liu2018image}.
  }
  \label{fig:compare}
\end{figure}
\begin{figure}
  \centering
  \includegraphics[width=6cm]{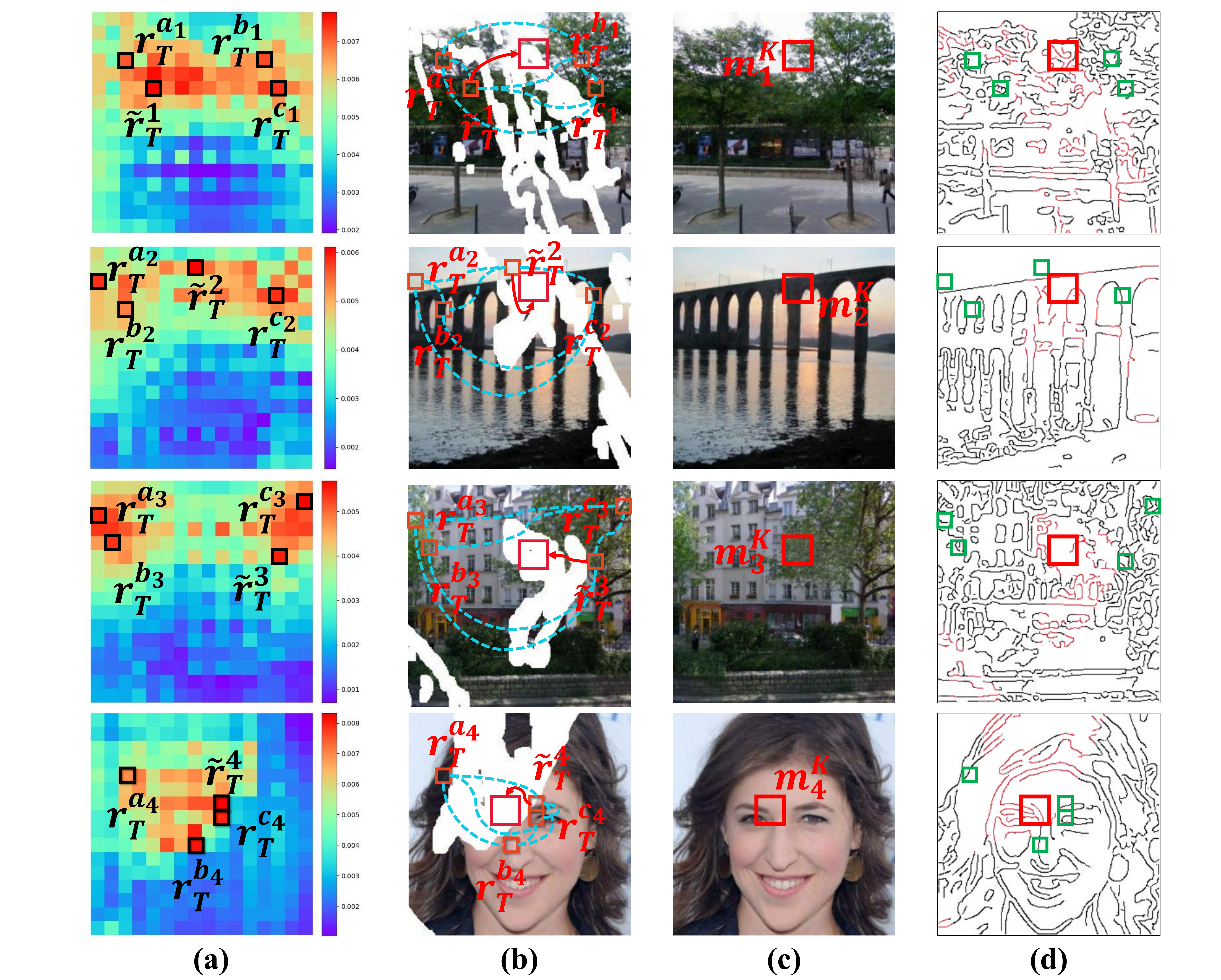}
  \caption{(a) show the attention maps of exemplar texture references for coarsely filled information over masked region; (b) exhibit the exemplar texture reference vectors to inpaint the masked patches; (c) and (d) visualize the inpainting output together with corresponding structure information.}
  \label{fig:relate}
\end{figure}

\begin{figure}
  \centering
  \includegraphics[width=6.5cm]{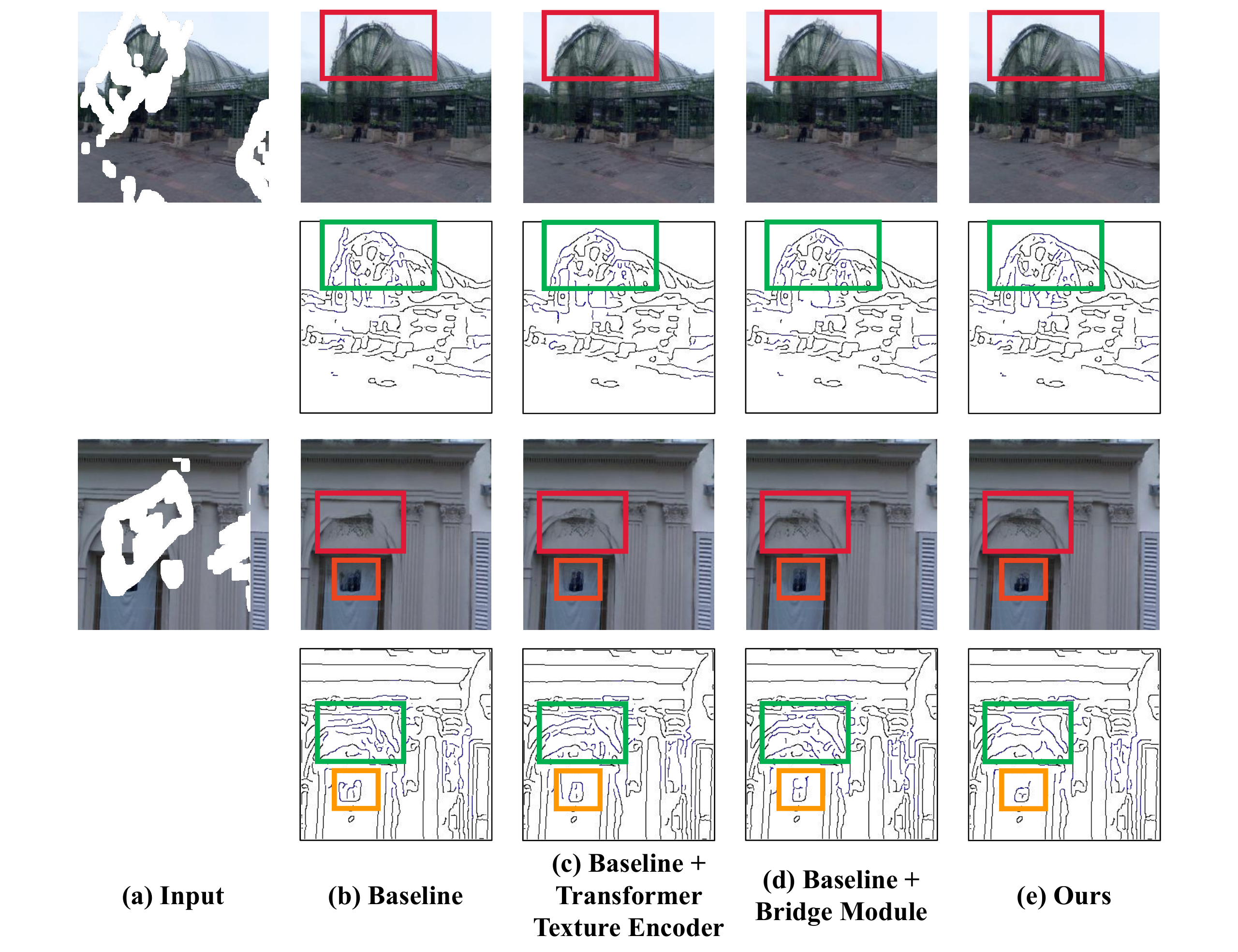}
  \caption{(a) are the input masked images; (b) are the baseline model; (c) and (d) are the varied models based on (b), Compared with (c) and (d), our method (e) achieves the better results by addressing the artifacts and synthesize plausible semantics on the patches marked by boxes.}
  \label{fig:ab}
\end{figure}

\section{Experiment}
\subsection{Implementation Details}
Our proposed method is implemented in Python and Pytorch. Following \cite{carion2020end}, we train architecture with AdamW \cite{loshchilov2018decoupled} optimizer; the learning rate of transformer and backbone set to $10^{-4}$ and $10^{-5}$ respectively, weight decay to $10^{-4}$. All transformer weights are initialized with Xavier init \cite{glorot2010understanding}, and we adopt the ImageNet-pretrained ResNet50 \cite{he2016deep} from TORCHVISION with frozen batchnorm layers. We also increase the feature resolution via a dilation to the last stage of the backbone and removing a stride from the first convolution of this stage.  Both the transformer encoder and decoder include four layers. The network is trained using $256 \times 256$ images with irregular masks \cite{liu2018image}, we conduct experiments on three public datasets that have different characteristics: Paris StreetView (PSV) \cite{doersch2012makes}, CelebA-HQ \cite{liu2015deep} and Places2  \cite{zhou2017places}. We use 2 NVIDIA 2080TI GPU with batch size 32 for PSV, 4 NVIDIA 2080TI GPU for CelebA-HQ and Places2 with batch size 64 to train the transformers.
\subsection{Comparison with state-of-the art methods}
We quantitatively evaluate our proposed method and state-of-the-arts as per four evaluation metrics: 1) ${\ell_{1}}$ error; 2)  peak signal-to-noise ratio (PSNR); 3) structural similarity index (SSIM) \cite{wang2004image}; and 4) Fréchet Inception Score (FID) \cite{heusel2017gans}. The ${\ell_{1}}$, PSNR and SSIM are used to compare the low-level differences over pixel level between the generated image and ground truth. FID evaluates the perceptual quality by measuring the feature distribution distance between the synthesized and real images. The irregular masked regions of the image are testified with varied ratios over the whole image size.\\
\textbf{Quantitative Comparisons.} We compare our method with the latest competitors\footnote{We only choose the competitors that report the quantitative results over the above three data sets with varied mask ratio.}: 1) CNNs texture-only methods: GC \cite{yu2019free} and PIC \cite{zheng2019pluralistic}; 2) Attention-based method: HiFill \cite{yi2020contextual}; 3) Structure-Texture based methods: MEDFE \cite{Liu2019MEDFE},  EC \cite{Nazeri_2019_ICCV}, CTSDG \cite{Guo_2021_ICCV}, EII \cite{wang2021image} and 4) decoder-only transformer based methods: ICT \cite{wan2021high}, BAT \cite{yu2021diverse}. For fairness, we directly report the comparable results from \cite{yu2021diverse,wang2021image,Guo_2021_ICCV}. As shown in Table. \ref{table:compare}, we can see that our method enjoys a smaller ${\ell_{1}}$ error and FID score, together with larger PSNR and SSIM than the competitors. Particularly, \textit{small FID score validates the pros for the global texture references and structure feature representation}. As the texture-only method GC and PIC only fill the masked patches via the bounded known regions. While HiFill independently calculates similarities between each coarsely filled patch and all known patches, which misleads the masked patch to be inpainted by only one dominant known region. Despite the intuition of  MEDFE to capture the structure information, it failed to exploit the information from all known patches. The similar limitation also holds for EC, CTSDG and EII, which keep consistent with our analysis in Sec. \ref{decoder}.  BAT and ICT restore the masked regions upon the pixel level, which cannot well capture the global texture semantics, in line with the principle in Sec. \ref{Encoder}. Our method outperforms the others.
\\
\textbf{Qualitative Comparisons.} To shed more light on the observation, Fig.\ref{fig:compare} showcases the visualization results over all methods over three datasets. It can be seen that the inpainting output by our method is more semantically coherent based on surrounding known regions. We validate such intuition by Fig.\ref{fig:relate}, where \ref{fig:relate}(a) show the examples of the texture reference vectors encoding global semantic correlations to be reconstructed via other texture references beyond the local regions, to yield a better inpainting output in \ref{fig:relate}(c). We further visualize the structure information of inpainting output in \ref{fig:relate}(d), which validate the intuition of our structure-texture matching attention in Sec. \ref{decoder}.  
\\
\textbf{User Study.} We further perform subjective user study over dataset PSV, CelebA-HQ and Places2. Concretely, we randomly sampled 20 test images from each dataset, Total 10 volunteers are invited to choose the most realistic images from inpainting results generated by the proposed method and some state-of-the-art approaches. As shown in the last column of Table. \ref{table:compare}, the result of our method outperforms state-of-the-arts by large margins.

\subsection{Ablation Study} To further validate the advantage of each component for our pipeline, we conduct the ablation studies over the following varied models:
\textbf{Baseline model.} The baseline model contains neither transformer texture encoder nor the bridge module formulated as Eq.\ref{eq:Decoder1}, which is equivalent to transformer decoder only model.  Unlike the existing models over pixel level, we perform such baseline model over patch level and  reconstruct the masked regions. Comparing Fig.\ref{fig:ab}(b) and (e), the baseline model inattentively produces the blurry noise in the restoring images, especially around complex boundaries such as the green architecture in the red box.  Table.\ref{table:ab} reports the quantitative results, which further discloses the contribution of  global texture reference and bridge module to the performance gain.\\
\textbf{The importance of  bridge module.} We abandon Eq.\ref{eq:Decoder1} for bridge module, and leave Eq.\ref{eq:Decoder} only to validate the importance of our bridge module. Our proposed attention transition manner aims at well matching the \textit{global texture} and \textit{structure} for inpainting. Comparing Fig.\ref{fig:ab}(c)  with (e),  the bridge module can obviously benefit the restoration of structure guided by the consistent global textures. Table. \ref{table:ab} also validates such observation.\\
\textbf{The importance of global texture reference.}  We testify the model that can only restore the masked regions with \textit{local texture} representation, \textit{i.e.,} obtaining texture representation via ResNet50 \textit{yet} without self-attention to yield texture references.   Fig.\ref{fig:ab}(d) offers the visualized results of such model,  comparing with our method in \ref{fig:ab}(e) together with the results in Table. \ref{table:ab}, which confirm the benefits of  our global texture reference to foster the inpainting performance.



\section{Conclusion}
In this paper, we delve globally into texture and structure information for image inpainting. Technically, the transformer pipeline paired with both encoder and decoder is proposed. The encoder aims at capturing the global texture semantic correlations across the whole image, while the decoder module restores the masked patches, featured with a structure-texture matching attention module to well match the global texture and structure information. An adaptive patch vocabulary is established, to progressively inpaint all the coarsely filled patches via a probabilistic diffusion process. Experimental results over benchmarks validate the advantages of our model over state-of-the-art counterparts.

\\
\textbf{Acknowledgments}
This work was supported by the National Natural Science Foundation of China under Grant No U21A20470, 62172136, 61725203, U1936217. Key Research and Technology Development Projects of Anhui Province (no. 202004a5020043).

\bibliographystyle{ACM-Reference-Format}
\balance
\bibliography{sample-base}

\end{document}